\documentclass{article} 
\usepackage[preprint]{colm2026_conference}

\usepackage{microtype}
\usepackage{hyperref}
\usepackage{url}
\usepackage{booktabs}

\usepackage{lineno}

\definecolor{darkblue}{rgb}{0, 0, 0.5}
\hypersetup{colorlinks=true, citecolor=darkblue, linkcolor=darkblue, urlcolor=darkblue}

\usepackage[most]{tcolorbox}
\newtcolorbox{findingbox}[1]{
  colback=gray!5,      
  colframe=black!75,   
  fonttitle=\bfseries,
  title=#1,            
  enhanced,
  sharp corners,       
  boxrule=0.5pt,
  left=5pt, right=5pt, top=5pt, bottom=5pt
}
\usepackage{svg}

\title{Rethinking Inference-Time Scaling in Local Computer-Use Agents: Failure Modes and Compute Tradeoffs}

\author{
\normalfont
\begin{tabular}{ll}
\textbf{Woongkyu Lee}
&
\textbf{Jungwook Choi \thanks {Corresponding author.}}
\\
\multicolumn{2}{l}{Hanyang University}
\\
\multicolumn{2}{l}{
\texttt{\{lwghanyang, choij\}@hanyang.ac.kr}
}
\end{tabular}
}

\begin{document}

\ifcolmsubmission
\linenumbers
\fi

\maketitle

\begin{abstract}
Deploying autonomous computer-use agents (CUAs) locally is increasingly important for privacy, cost efficiency, and practical usability, yet improving their performance under strict hardware constraints remains challenging. While recent studies show that inference-time scaling can improve frontier computer-use agents through additional computation during execution, its effectiveness for resource-constrained local models remains poorly understood. We present a systematic empirical study of inference-time scaling in local CUAs across contextual, temporal, structural, and parallel dimensions. We evaluate Qwen3-VL-8B/30B-A3B, UI-TARS-1.5-7B, and OpenCUA-7B on the OSWorld benchmark. Our results show that additional computation often yields diminishing returns while changing failure modes. Contextual scaling provides historical grounding that improves trajectory stability and task accuracy, but its gains saturate as token cost increases and failures shift from repetitive or stalled trajectories toward premature false successes. Temporal scaling similarly reduces max-step stalls, yet does not substantially improve task success, indicating that longer horizons often extend erroneous trajectories rather than correct them. We further find that structural decomposition can introduce planning and formatting overhead in local two-stage agents, while parallel scaling partially mitigates these failures at a substantial computational cost. Overall, our findings suggest that efficient local CUAs require selective compute allocation, failure-aware control mechanisms, and agentic frameworks designed around the capabilities and limitations of local models.

\end{abstract}

\section{Introduction}

The emergence of Large Language Models (LLMs) has fundamentally shifted the paradigm of digital interaction, moving beyond simple text generation toward autonomous Computer-use Agents (CUAs) capable of navigating complex Graphical User Interfaces (GUIs). 
Unlike traditional task-specific bots, these agents leverage multimodal reasoning to perceive screen states and interaction history, execute GUI actions (e.g., click, type, and scroll) to fulfill high-level user instructions. 
This progress has been fueled by both the development of realistic evaluation benchmarks~\citep{xie2024osworldbenchmarkingmultimodalagents, deng2023mind2webgeneralistagentweb, bonatti2024windowsagentarenaevaluating, zhou2024webarenarealisticwebenvironment} and the emergence of specialized multimodal models~\citep{qin2025uitarspioneeringautomatedgui, wang2025opencuaopenfoundationscomputeruse, xue2026evocuaevolvingcomputeruse, lin2024showuivisionlanguageactionmodelgui} designed for GUI interaction and action grounding. Together, these advances have enabled rapid progress in building agents capable of operating in complex, real-world digital environments.

Recent advancements have introduced highly capable agentic architectures. For instance, the Agent S series~\citep{agashe2024agentsopenagentic, agashe2025agents2compositionalgeneralistspecialist, gonzalezpumariega2026scalingagentscomputeruse} demonstrated the effectiveness of hierarchical planning and sub-goal management, while JEDI~\citep{xie2025scalingcomputerusegroundinguser} and GTA-1~\citep{yang2025gta1guitesttimescaling} adopt decoupled planning–grounding frameworks, where high-level plans are generated by a reasoning model and executed by a specialized grounding model. To further improve performance, recent work explores inference-time scaling, which allocates additional computation during execution to enhance agent accuracy~\citep{gonzalezpumariega2026scalingagentscomputeruse, yang2025gta1guitesttimescaling}. In computer-use agents, this is realized through techniques such as trajectory-level scaling and parallel plan generation, leading to notable gains in task success.

However, a fundamental tension remains between model capability and operational efficiency. While state-of-the-art agents achieve strong performance using proprietary models \citep{singh2025openaigpt5card, sonnet4.6}, they incur substantial costs in step counts and token consumption compared to human efficiency \citep{abhyankar2025osworldhumanbenchmarkingefficiencycomputeruse}. Local CUAs based on small open-source models are therefore important for practical and secure deployment. Yet, although inference-time scaling~\citep{yang2025gta1guitesttimescaling,gonzalezpumariega2026scalingagentscomputeruse} improves large proprietary agents, it remains unclear whether the added computation is worthwhile for resource-constrained local models, or which failure patterns limit further gains. This motivates a systematic study of the marginal utility and failure behavior of inference-time scaling in local settings.


In this paper, we evaluate the inference-time scaling behaviors of local CUAs, including Qwen3-VL, UI-TARS, and OpenCUA, on the OSWorld benchmark. To understand how inference-time scaling affects both efficiency and failure behavior, we categorize it into \textit{contextual}, \textit{temporal}, \textit{structural}, and \textit{parallel} dimensions, analyzing how each contributes to task success. Our empirical analysis reveals that the scaling of local CUAs is governed by non-linear dynamics, where incremental resource allocation often leads to diminishing returns. Specifically, while minimal history is a prerequisite for stability, excessive contextual scaling and structural decomposition trigger saturation effects and reasoning bottlenecks that escalate costs without accuracy gains. These observations suggest that inference-time scaling in local CUAs changes failure modes more than it increases task success, requiring compute to be allocated selectively around the limitations of local models.

The primary contributions of this work are as follows:
\begin{itemize}
\item \textbf{A Systematic Study of Local CUA Scaling:}
We analyze inference-time scaling in local GUI agents across \textbf{contextual scaling} (history length), \textbf{temporal scaling} (maximum steps), \textbf{structural decomposition} (single-agent vs. two-stage agents), and \textbf{parallel scaling} (multiple plans). This provides a unified empirical view of how additional computation affects task success and cost in local CUAs.

\item \textbf{Failure-mode Analysis of Scaling Behavior:}
We show that additional computation often changes failure patterns rather than improving task success. In particular, longer histories and larger step budgets reduce repetitive loops and max-step stalls, but increasingly shift failures toward premature false successes. We also identify planning and format-parsing failures as key limitations of two-stage local agents.
\item \textbf{Practical Directions for Efficient Local CUAs:}
Based on the observed accuracy–cost tradeoffs and failure patterns, we derive practical directions for efficient local CUA deployment. Our analysis suggests using moderate history lengths, bounded temporal budgets, structurally simple agents, and selective parallel scaling. It further motivates future designs that combine context management, failure-aware control, and agentic frameworks aligned with local model capabilities.
\end{itemize}

The remainder of this paper is structured to lead from empirical observation to practical optimization. Sections 2 and 3 establish the research context and methodology. Section 4 presents our core analysis, beginning with the scaling dynamics of single agents and the structural overhead of two-stage architectures, and examines how additional computation affects both task success and failure patterns. Section 5 discusses the implications of these findings, including resource allocation strategies, failure-aware control, and design principles for efficient local CUAs. Finally, Section 6 concludes the paper and outlines future research directions.

\section{Related Work}

\textbf{Autonomous Computer-Use Agents} The development of autonomous agents capable of navigating Graphical User Interfaces (GUIs) has been accelerated by the introduction of comprehensive benchmarks\citep{xie2024osworldbenchmarkingmultimodalagents, bonatti2024windowsagentarenaevaluating, deng2023mind2webgeneralistagentweb, zhou2024webarenarealisticwebenvironment}. Existing approaches can be broadly categorized into two directions. 
The first direction leverages single-model architectures~\citep{sonnet4.6, bai2025qwen3vltechnicalreport, wang2025opencuaopenfoundationscomputeruse, qin2025uitarspioneeringautomatedgui}, where a multimodal model directly maps visual observations and task instructions to executable actions in an end-to-end manner. The second direction adopts agentic frameworks that decompose the decision-making process into multiple stages, such as planning and grounding. Frameworks such as Agent S~\citep{agashe2024agentsopenagentic, agashe2025agents2compositionalgeneralistspecialist, gonzalezpumariega2026scalingagentscomputeruse} and GTA-1~\citep{yang2025gta1guitesttimescaling} have demonstrated that structured decomposition can improve task success through hierarchical planning and coordination. However, despite their effectiveness, these agents rely heavily on proprietary API-based models for planning, motivating the study of local CUAs based on open-source models under strict hardware and inference-time constraints.

\textbf{Inference-time Scaling Laws} Inference-time scaling, defined as improving performance by allocating additional computational resources during execution, has gained prominence following the success of large-scale reasoning models~\citep{snell2024scalingllmtesttimecompute, wang2023selfconsistencyimproveschainthought, yao2023treethoughtsdeliberateproblem}.  Recent work has begun to explore this paradigm in the context of computer-use agents. For instance, Agent S3 introduces trajectory-level scaling via multiple rollouts and behavior selection~\citep{gonzalezpumariega2026scalingagentscomputeruse}, while GTA-1 employs parallel test-time scaling through multiple candidate plans~\citep{yang2025gta1guitesttimescaling}. Despite these advances, the empirical relationship between inference-time parameters and task success in multimodal GUI environments remains under-explored, which we address by systematically analyzing the interplay between the scaling dimensions.

\textbf{Efficiency and Context Management in CUAs} 
Operating computer-use agents in high-density GUI environments requires processing high-resolution screenshots and interaction histories, leading to substantial computational overhead. Recent work emphasizes efficiency in evaluation, with OSWorld-Human introducing step-based human baselines and revealing significant efficiency gaps~\citep{abhyankar2025osworldhumanbenchmarkingefficiencycomputeruse}. To mitigate long-context costs, prior approaches explore compression techniques, including key-value cache compression and token pruning~\citep{huang2025guikvefficientguiagents, chen2025moreempoweringguiagent}, as well as summarizing visual histories into compact textual representations~\citep{liu2025paluiplanningactivelookback}. 
However, the tradeoff between maintaining sufficient history for task continuity and minimizing computational cost remains under-explored for local-scale agents, which we address by analyzing efficiency-performance tradeoffs across history usage.

\section{Methodology}

This section describes the experimental design for evaluating the accuracy-cost tradeoff of local computer-use agents according to scaling parameters.

\subsection{Structure of CUA}

\begin{figure}[t]
\centering
\includegraphics[width=\linewidth]{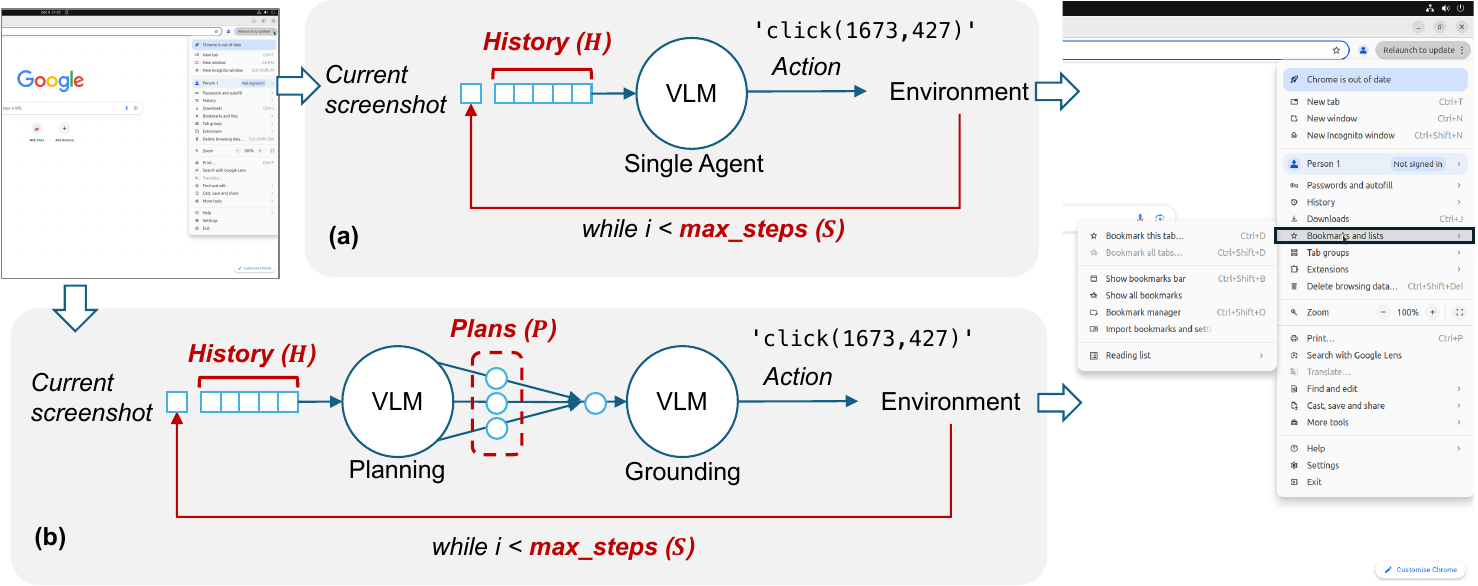}
\caption{Illustration of CUA frameworks and inference-time scaling dimensions. (a) The Single Agent directly maps the current screenshot and history length ($H$) to an action within the maximum temporal budget ($S$). (b) The Two-stage Agent decouples this process into a Planning stage that generates multiple parallel candidate plans ($P$) and a Grounding stage that executes the selected action.}
\label{fig:framework}
\end{figure}

To improve the accuracy of computer-use agents, various agentic frameworks have been proposed. We categorize these into two representative structures, illustrated in Fig.~\ref{fig:framework}, and conduct experiments accordingly:

\begin{enumerate}
\item \textbf{Single Agent (Fig.~\ref{fig:framework}a)}: This is the basic structure for a computer-use agent. A single model receives the task instruction, the current screenshot, and a history of previous actions and screenshots ($H$) as input to generate an action at each step \citep{qin2025uitarspioneeringautomatedgui, wang2025opencuaopenfoundationscomputeruse}. This process iterates in a loop until the agent reaches the task goal or the maximum decoding steps ($S$) are exhausted.
\item \textbf{Two-stage Agent (Fig.~\ref{fig:framework}b)}: This structure decouples inference into \textit{Planning} and \textit{Grounding} stages \citep{xie2025scalingcomputerusegroundinguser, yang2025gta1guitesttimescaling}. A planner receives the current screenshot and history ($H$) to generate multiple candidate plans ($P$). A judging process then selects the best plan, which a specialized grounding model translates into a final action with specific coordinates. This process iterates until the goal is met or the maximum steps ($S$) are reached. While more complex structures incorporating coding agents \citep{song2026coact1computerusingmultiagentcoding} or memory modules \citep{agashe2024agentsopenagentic} exist, they fundamentally follow this two-stage paradigm.
\end{enumerate}

\subsection{Compute Scaling Dimensions}

Computer-use agents allow for scaling across various dimensions during inference-time. We identify three compute scaling factors within each agent architecture (highlighted in red in Fig.~\ref{fig:framework}), and examine their impact on accuracy and cost.

\subsubsection{Temporal Scaling: Max Decoding Steps ($S$)}The temporal budget for task completion is inherently limited and is typically enforced through a maximum step constraint. Each step corresponds to a single agent–environment interaction, where the agent observes the current state and generates an action. As $S$ increases, the agent is afforded more opportunities to reach a solution. We measure performance by scaling the maximum steps to 15, 50, and 100, following the OSWorld benchmark standards.

\subsubsection{Contextual Scaling: History Image Length ($H$)}Agents can utilize previous actions and screenshot history as context to maintain continuity. While history can improve accuracy, the resulting increase in token usage is inevitable. Most modern computer-use agents leverage Multimodal Large Language Models (MLLMs), where high-resolution screenshot histories lead to significant visual token consumption \citep{huang2025guikvefficientguiagents, liu2025paluiplanningactivelookback}. In this paper, we scale the number of screenshot histories to $H \in \{0, 1, 4, 8\}$ to analyze the resulting tradeoffs.

\subsubsection{Parallel Scaling: Number of Multiple Plans ($P$)}Parallel test-time scaling involves generating multiple outputs to improve agent accuracy. For instance, a planning agent in a two-stage agent may generate multiple plans, or multiple agents may run in parallel to select the optimal trajectory. Following the approach of GTA-1~\citep{yang2025gta1guitesttimescaling}, we investigate the effects of varying the number of high-level plans ($P$) in a two-stage agent for local CUAs.

\subsection{Cost Metrics}
To evaluate the operational cost in each setting, we utilize the \textbf{number of steps per task} and \textbf{prompt token usage} as primary metrics. The number of steps measures how many actions an agent takes on average to complete a task, serving as a direct indicator of execution efficiency~\citep{abhyankar2025osworldhumanbenchmarkingefficiencycomputeruse}. Prompt token usage is calculated as the cumulative number of tokens processed during all model calls within a single task, which reflects the dominant computational cost in GUI-based interaction, especially because repeated high-resolution screenshot histories dominate the input cost across agent architectures.



\subsection{Experimental Setup}

\textbf{Benchmark.} We utilize OSWorld~\citep{xie2024osworldbenchmarkingmultimodalagents}, a representative benchmark in the computer-use domain. Our experiments were conducted on 361 real-world Ubuntu tasks, excluding eight Google Drive tasks due to environment constraints. Agents operate solely from screenshots, without access to auxiliary interfaces such as accessibility (a11y) trees or visual prompting techniques like Set-of-Mark (SoM)~\citep{yang2023setofmarkpromptingunleashesextraordinary}.

\textbf{Models.} For the single agent framework, we evaluated four models: Qwen3-VL-8B-Instruct, Qwen3-VL-30B-A3B-Instruct (MoE)~\citep{bai2025qwen3vltechnicalreport}, UI-TARS-1.5-7B~\citep{qin2025uitarspioneeringautomatedgui}, and OpenCUA-7B~\citep{wang2025opencuaopenfoundationscomputeruse}. For the two-stage framework, the grounding agent was fixed as GTA-1-7B~\citep{yang2025gta1guitesttimescaling}. The planning and judging roles were performed by the two Qwen3 models (8B and 30B-A3B).

\textbf{System.} All experiments were conducted on an A100-80GB GPU. We utilized the vLLM engine for model inference and to measure precise token usage for each agentic configuration.
\section{Results}

\begin{figure}[t]
\centering
\includesvg[width=\linewidth]{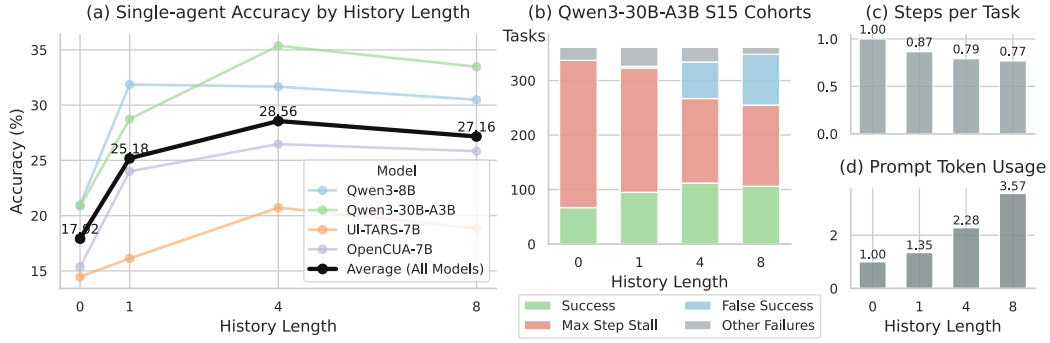}
\caption{Impact of history length on single-agent performance, efficiency, and failure modes. 
(a) Task success rates across different local models. 
(b) Failure-mode cohort analysis of Qwen3-VL-30B-A3B ($S=15$), showing that increasing history reduces max-step stalls but shifts failures toward false successes, leading to performance saturation beyond $H=4$.
(c) Normalized average steps per task. 
(d) Normalized prompt token usage per task. 
}
\label{fig:single_hist}
\end{figure}

\subsection{The Dynamics of Contextual and Temporal Scaling (Single Agent)}


\begin{findingbox}{Finding 1: Minimal history is a fundamental prerequisite for local agent stability.}
Our empirical results reveal that providing even a single previous screenshot ($H=1$) leads to a substantial performance leap across all evaluated models. Expanding to $H=4$ further improves the context for tracking complex state transitions.
\end{findingbox}

Fig.~\ref{fig:single_hist} (a) illustrates the impact of the contextual scaling on the success rates of single-pass local agents. On average, accuracy surges from approximately 18\% at $H=0$ to over 25\% at $H=1$. This sharp gain suggests that minimal history is essential for stabilizing action sequences. Without access to recent history, local agents frequently fall into repetitive loops or execute redundant actions, as they lack the context to verify whether a previous instruction was successfully carried out.

\begin{figure}[t]
\centering
\includegraphics[width=\linewidth]{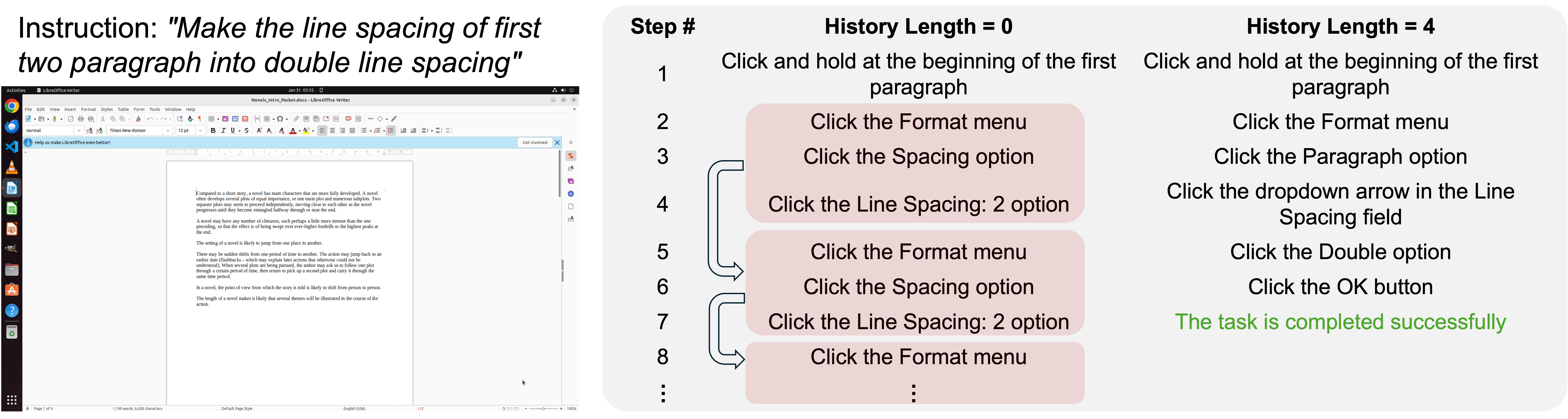}
\caption{Comparison of agent trajectories by contextual scaling. Without contextual grounding ($H=0$), the agent falls into a repetitive loop. With adequate history ($H=4$), the agent leverages previous states to successfully complete the multi-step task.}
\label{fig:hist}
\end{figure}

As shown in the qualitative example in Fig.~\ref{fig:hist}, an agent with $H=0$ fails to progress by repeatedly clicking the same menu items, effectively becoming trapped in an execution loop. In contrast, providing adequate history ($H=4$) allows the agent to recognize previous states and successfully navigate the multi-step sequence to completion. While $H=1$ serves as the primary stabilizer, expanding the history to $H=4$ provides the necessary context for tracking complex state transitions in long-horizon tasks. However, our data indicates that this upward trend does not continue linearly, pointing toward a specific threshold where additional context no longer translates to task success.

\begin{findingbox}{Finding 2: History improves step efficiency but shifts failures toward false successes.}
Increasing history length reduces redundant actions and lowers max-step stalls, but accuracy saturates beyond $H=4$ as premature false successes increase. This indicates that additional history improves stability but can also lead to incorrect termination.
\end{findingbox}

The failure-mode analysis in Fig.\ref{fig:single_hist} (b) shows that increasing $H$ reduces max-step stalls, suggesting that historical context helps agents avoid repetitive or stalled trajectories.
This stabilization is also reflected in Fig.\ref{fig:single_hist} (c), where longer histories reduce the average number of steps per task.
However, simply extending history does not eliminate failures.
Instead, the dominant failure mode shifts toward premature false successes, where the agent terminates despite not completing the task.
These observations suggest that contextual scaling primarily improves trajectory stability, while its effect on task-completion judgment remains limited.
As a result, $H=4$ (28.56\%) offers the best observed tradeoff, while increasing history to $H=8$ (27.16\%) adds cost without improving accuracy.

\begin{figure}[t]
\centering
\includesvg[width=\linewidth]{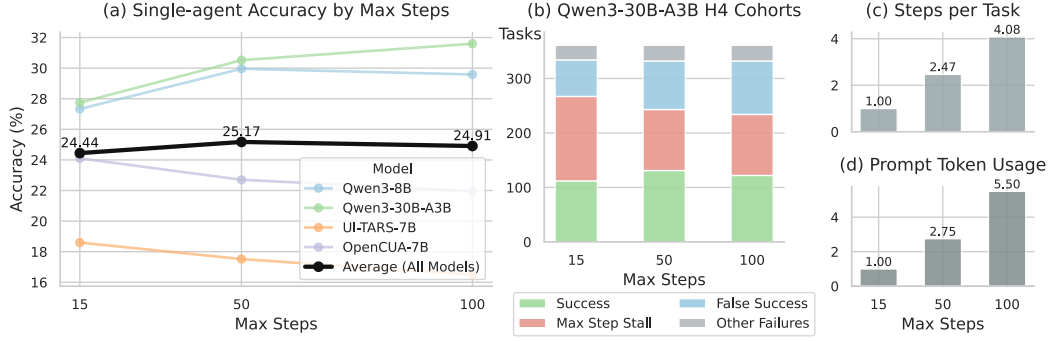}
\caption{
Impact of temporal scaling on single-agent. 
(a) Task success rates across different local models. 
(b) Normalized average steps per task. 
(c) Normalized prompt token usage per task. 
(d) Failure-mode cohorts for Qwen3-VL-30B-A3B ($H=4$). Additional steps shift failures from max-step stalls to premature false successes rather than increasing task success.
}
\label{fig:single_steps}
\end{figure}

\begin{findingbox}{Finding 3: Temporal scaling is ineffective under limited reasoning capacity.}

Extending decoding steps ($S$) fails to overcome the reasoning limitations of local models. While accuracy remains largely unchanged, operational costs grow linearly. Additional steps also shift failures from max-step stalls to premature false successes, making temporal scaling inefficient for improving task success in resource-constrained settings.
\end{findingbox}

We evaluate temporal scaling by varying the maximum number of steps ($S$). As shown in Fig.~\ref{fig:single_steps} (a), increasing $S$ from 15 to 100 yields little to no improvement in task success across most models. Although Qwen3 shows some improvement when increasing $S$ from 15 to 50, the gains remain marginal beyond this range.

Fig.\ref{fig:single_steps} (b) shows that increasing $S$ reduces max-step stalls but shifts failures toward premature false successes.
This suggests that additional steps help agents avoid early termination by step limits, but do not necessarily help them recover from incorrect trajectories.
Instead, longer horizons often allow agents to continue erroneous behavior.
This failure-mode redistribution explains why task success remains largely unchanged despite larger step budgets.
At the same time, operational cost increases steadily with $S$, as reflected by both average steps per task and prompt token usage (Fig.\ref{fig:single_steps} (c), (d)).
Thus, temporal scaling increases cost while primarily redistributing failures rather than raising the success ceiling.

\subsection{Structural Decomposition and Parallel Scaling (Two-stage Agent)}

\begin{findingbox}{Finding 4: Structural decomposition can introduce overhead in local-scale agents.}
In our experiments, decoupling reasoning into planning and grounding stages reduced task success while increasing computational cost. Compared to single-pass agents, two-stage agents incur additional planning and formatting requirements that introduce new failure modes.
\end{findingbox}

\begin{figure}[t]
\centering
\includesvg[width=\linewidth]{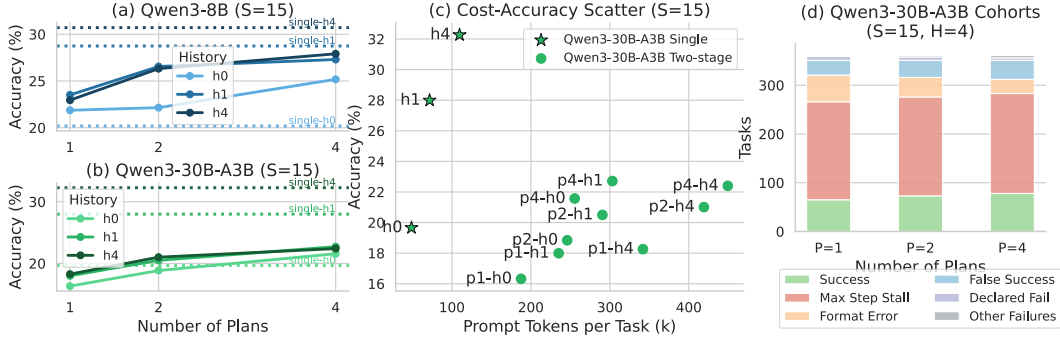}
\caption{
Impact of parallel scaling on two-stage agent. 
(a,b) Task success rates of Qwen3-8B and Qwen3-30B-A3B planners across different numbers of plans ($P$) and history lengths ($H$) with $S=15$. Dotted lines indicate the corresponding single-agent performance.
(c) Accuracy--cost tradeoff, where single-agent configurations (stars) achieve higher accuracy at substantially lower token cost than two-stage configurations (circles).
(d) Failure-mode cohorts for the Qwen3-VL-30B-A3B two-stage agent ($S=15$, $H=4$).
}
\label{fig:two}
\end{figure}

As shown in Fig.~\ref{fig:two} (a,b), the two-stage framework consistently underperforms the corresponding single-agent baselines across history lengths. At the baseline configuration ($P=1$), task success drops substantially despite the additional planning stage, indicating that structural decomposition alone does not improve performance in local settings.

\begin{figure}[t]
\centering
\includegraphics[width=\linewidth]{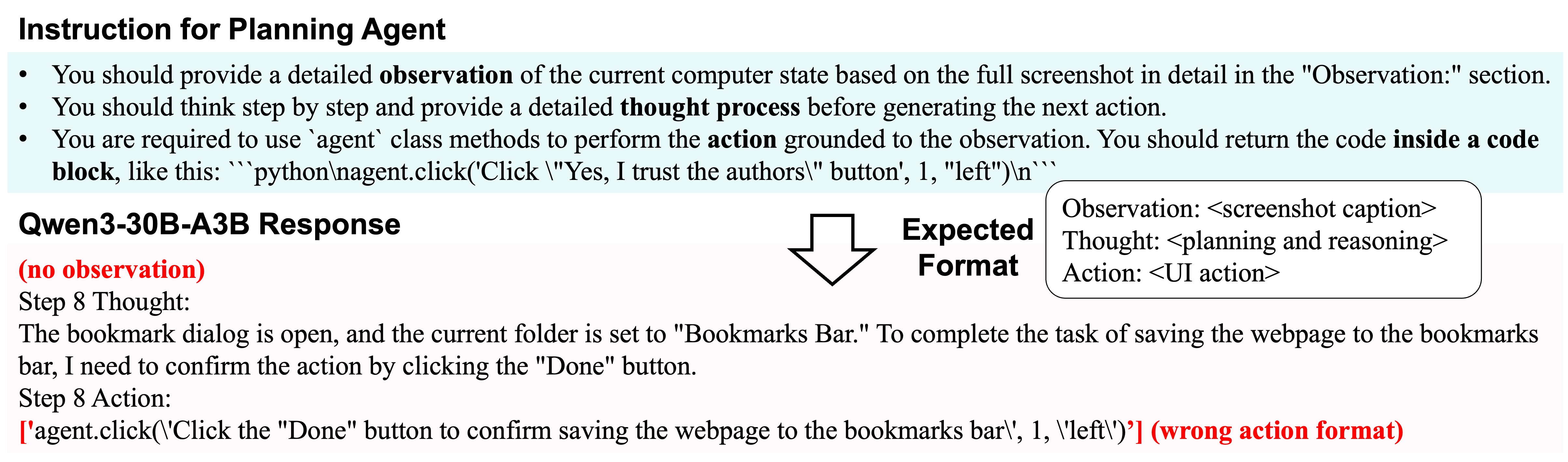}
\caption{
Example of a planner format error. Despite explicit formatting instructions, the planner omits the observation section and generates an action as a plain string rather than a valid code block, resulting in a parsing failure.
}
\label{fig:format-error}
\end{figure}

The cost--accuracy tradeoff in Fig.~\ref{fig:two} (c) further highlights this overhead. 
Two-stage configurations consume substantially more tokens while achieving lower accuracy than single agents. 
This may stem from limited planning capability in local models, where generated plans can be incomplete or poorly aligned with the visual state. Moreover, decomposition introduces a formatting dependency between planning and grounding. As illustrated in Fig.~\ref{fig:format-error}, the planner can omit required fields or generate actions outside the required code block. Fig.~\ref{fig:two} (d) shows that format errors account for a non-trivial portion of failures, preventing successful parsing and execution by the grounding agent.

\begin{findingbox}{Finding 5: Parallel scaling partially recovers performance but only with sub-linear gains relative to computation.}
Increasing the number of candidate plans partially offsets the weaknesses of two-stage decomposition by improving plan selection and reducing format-related failures. However, this recovery requires substantially higher token consumption, yielding a clear accuracy--cost tradeoff.
\end{findingbox}

While structural decomposition suppresses performance at baseline, increasing the number of parallel rollouts ($P$) partially recovers accuracy. As shown in Fig.~\ref{fig:two} (a,b), performance improves as $P$ increases from 1 to 4, suggesting that multiple candidate plans provide additional opportunities to select an executable and better-aligned action.

Fig.~\ref{fig:two} (d) further shows that parallel scaling reduces format errors. 
Because the planner generates multiple candidate plans, invalidly formatted outputs can be bypassed when at least one valid candidate remains available for grounding. This helps mitigate one failure mode introduced by decomposition, although it does not fully close the gap with single-agent baselines.

This improvement, however, comes at a significant cost. Token usage increases sharply with $P$, reflecting the overhead of generating multiple plans (Fig.~\ref{fig:two} (c)). Overall, parallel scaling introduces a clear tradeoff: it partially recovers task success and reduces format-related failures, but the gains remain sub-linear relative to the increase in computation.

\section{Discussion}

\subsection{Resource Allocation Strategy and Practical Guidelines}


Based on the observed accuracy–cost tradeoffs and failure-mode patterns, we propose that compute in local CUAs should be allocated selectively rather than scaled uniformly.


\textbf{Contextual Scaling and Historical Grounding}
Contextual scaling offers the strongest initial benefit. Moving from $H=0$ to $H=1$ reduces repetitive loops and max-step stalls, while $H=4$ further improves step efficiency. However, longer histories increasingly shift failures toward premature false successes, suggesting that moderate history is preferable to unbounded context expansion.

\textbf{Limitations of Temporal Scaling}
Temporal scaling provides limited gains. Increasing $S$ reduces max-step stalls but does not substantially improve task success; instead, failures often shift toward premature false successes. Thus, $S$ should be treated as a bounded budget aligned with the model’s reasoning capacity.

\textbf{Strategic Use of Parallel Scaling}
Parallel scaling can partially recover two-stage performance by increasing plan diversity and reducing format-failures. However, its gains remain sub-linear relative to the token cost, so it should be used selectively rather than by default.

Overall, we recommend prioritizing structural simplicity, calibrating temporal budgets, and using moderate contextual history to balance efficiency and trajectory stability.

\subsection{Toward Efficient Local Computer-Use Agents}

Our findings suggest that improving local CUAs requires targeted optimization rather than unconstrained inference-time scaling. Since additional compute often shifts failure modes instead of resolving them, future work should focus on context quality, failure-aware control, and model-aware agent design.


\textbf{Efficient Context Management}
Contextual scaling improves accuracy by providing historical grounding, but it also increases token cost and its benefit saturates when histories become too long. This suggests that local CUAs need mechanisms to retain task-relevant history while filtering out less useful past observations. Rather than simply expanding the context window, local agents should favor selective memory retention and compact history representations, as explored in PAL-UI \citep{liu2025paluiplanningactivelookback}.


\textbf{Failure-Aware Agent Control}
Our failure analysis shows that local agents often suffer from repetitive loops, stalled progress, and premature false successes. These patterns suggest that additional steps alone are insufficient unless agents can detect when they are no longer making meaningful progress or when they have terminated incorrectly. Event-driven cascades \citep{wei2026stepleveloptimizationefficientcomputeruse} address similar risks by escalating to stronger models when progress stalls or semantic drift is detected. However, fully local CUAs require lightweight alternatives to large-model recovery, including loop detection, trajectory revision, and task-completion verification.

\textbf{Agentic Frameworks for Local CUAs}
Our results also indicate that agentic frameworks designed for stronger models may not transfer directly to local models. Complex decomposition can introduce additional failures when local models cannot reliably follow the instructions. Therefore, local multi-agent systems should match the structure to model capability, or adapt each model to its assigned role through targeted training or fine-tuning.

\section{Conclusion}

We systematically studied inference-time scaling in local CUAs across contextual, temporal, structural, and parallel dimensions. Our results show that additional computation does not necessarily translate into higher task success. Contextual and temporal scaling can improve stability by reducing repetitive loops and max-step stalls, but their gains saturate as failures increasingly shift toward premature false successes. Structural decomposition can further introduce planning and format-parsing overhead in local two-stage agents, while parallel scaling partially mitigates these issues at a substantial computational cost. These findings suggest that future progress in local CUAs will depend less on increasing inference-time compute than on designing agents that allocate compute according to the capabilities and limitations of local models.

\bibliography{colm2026_conference}
\bibliographystyle{colm2026_conference}


\end{document}